# Semi-Supervised Learning Approach to Discover Enterprise User Insights from Feedback and Support

Xin Deng[1]     Ross Smith[1]     Genevieve Quintin[1]

[1]Microsoft, Redmond, WA, USA

xinde@microsoft.com, rosss@microsoft.com , gquintin@microsoft.com

*Abstract*—With the evolution of the cloud and the current customer centric culture, we inherently accumulate huge repositories of textual reviews, customer feedback, and support data. Whether it be to capture the customers sentiment of a store, a product, a specific feature, or satisfaction with the end to end experience, we have a large and growing set of data. This has driven enterprises to seek, produce and research engagement recognition patterns, user network analysis, topic detections, sentiment analysis etc. All with the goal to help prioritize product improvement and self-help efforts in the appropriate areas to reduce customer pain points. However, in many enterprises, huge manual work is still necessary to mine that data to be able to mine actionable outcomes. In this paper, we proposed and developed an innovative Semi-Supervised Learning approach by utilizing Deep Learning and Topic Modeling to have a better understanding of the customer's voice. This approach combines a BERT-based multiclassification algorithm through supervised learning combined with a novel Probabilistic and Semantic Hybrid Topic Inference (PSHTI) Model through unsupervised learning, aiming at automating the process of better identifying the main topics or areas as well as the sub-topics or specific issues from the textual feedback and support data. There are three major contributions and break-throughs from this work: 1). As the advancement of the deep learning technology progresses, there have been tremendous innovations in the Natural Language Processing (NLP) field, yet the traditional topic modeling and categorization as one of the NLP applications lag behind the tide of deep learning. In the methodology and technical perspective, we adopt transfer learning to fine-tune a BERT-based multiclassification system to categorize the main topics and then utilize the novel PSHTI model to infer the sub-topics under the predicted main topics. 2). The traditional unsupervised learning-based topic models or topic clustering methods suffer from the difficulty of automatically generating a meaningful topic label, but our system enables mapping the top words to the self-help issues by utilizing domain knowledge about the product through web-crawling.  3). This work provides a prominent showcase by leveraging the state-of-the-art methodology in the real production to help shed light to discover user insights and drive business investment priorities. When we think more deeply about where AI and machine learning can impact human efforts, topic modeling is an unparalled opportunity for machines to augment, rather than replace, human intelligence.

I. INTRODUCTION

*A. Deep Learning Evolves as Norm of Business in Natural Language Processing*

With the inspirational revolution of neural networks and deep learning, there have been continuous research efforts and achievements in Machine Learning, Computer Vision, Natural Language Processing (NLP), Speech Recognition, Robotics and much more. We are at the dawn of a new era where technology can augment the human experience. While the invention of the wheel or the printing press were impressive technological achievements that changed society, advances in machine learning are opening the door to a new world ahead. Bidirectional Encoder Representations from Transformers (BERT) is one of the greatest recent advances in the Natural Language Processing (NLP) field [1]. Moreover, the adoption of BERT and some advanced work built on BERT is becoming ubiquitous in a variety of NLP applications. BERT representations has been used in the medical field for understanding the clinical patient complaints [2]. Sentence embeddings were pre-trained through Siamese BERT-networks for the further applications like semantic similarity search [3]. These are exciting times where we can apply machine learning technology to new areas to deliver impactful results.

Consequentially, with the significant breakthroughs in the theoretical research, a variety of deep learning techniques including BERT also start to play an influential role in the real production and business industry. Around November 2019, Google announced their most recent major search update with the inclusion of BERT[1]. It helps better understand the intent behind users' search queries and return more relevant results.

As computing power has increased, technologies like BERT have been able to impact humanity in new and unique ways. In the domain of customer service and support, this research unlocks new and bold opportunities.

---

[1] Refer to https://searchengineland.com/faq-all-about-the-bert-algorithm-in-google-search-324193

*B. The Gaps Between Customer Service and the Rapid Development of Deep Learning Technology*

For many business and commercial companies, understanding customer behavior, as well as opinions or support requests regarding the products or services has been playing an essential role in providing insights into both quantity and quality perspectives. However, legacy systems based on keywords, simple clustering algorithms or traditional topic models are mostly adopted in production in the Customer Service teams within many enterprises. Facing the vast amount of textual reviews, customer feedback, and support data, the systems fail to automatically recognize the specific user intent or pain points or effectively identify the topics from the majority of the feedback or support requests. Consequently, thousands of Customer Support Engineers may spend hours talking to customers every single day, while other engineers have to look at large amount of feedback and support data in search of insights which inevitably requires huge manual work.

Mining detailed opinions and intents buried in the large amount of user feedback and support data remains an important but challenging task. Human tagging and mining is tedious and costly. Additionally, the sole fact of manual work entails the complexity of ensuring all the people mining the data have the exact same perception and judgment of the feedback. More intelligent and automated mechanisms become critically in need to serve such a purpose [4]. Replacing current legacy systems by adopting the advanced deep learning techniques can enable automating the process of identifying topics or problems from the user feedback and support data faster and more accurately.

Once the top topics and sub-topics are accurately identified, the industry can leverage the data points to make data driven prioritization decisions that are centered on the customers. For example, customer support teams could realize that they need to invest in automated diagnostics versus contextual self-help assets. Program managers could change their feature investment strategies to better align with customer pain points.

*C. The Contributions from the Work*

In the digital era, high volumes of the customer activity, feedback and support data accumulate exponentially and become available to the businesses. The massive data and powerful computational resources promote the development of various deep learning techniques, such as BERT as the representative of Attention Networks [1], Graph Neural Networks and so on. However, there has been a huge limitation due to the lack of labeled data. Companies have to invest huge amount of money to hire vendors for labeling the data. Solutions like self-supervised learning have been paid greater attention in both research and production. The two tasks Masked Word Prediction and Next Sentence Prediction adopted in BERT are self-supervised learning ways [1]. Most recently, Google expanded the exploration of self-supervised learning in text summarization as an objective and developed PEGASUS model [5]. Yet, in the customer support domain, especially regarding topic detection from the feedback and support data, both supervised learning based classification methods [6, 7] or unsupervised learning based methods [8,9] have been applied. This work proposes a semi-supervised learning approach consisted of two components, to leverage supervised learning which can make use of the existing available labeled main topics and unsupervised learning to identify sub-topics where we are lack of labeled data or ground truth. Fig 1 illustrates an example how this proposed new approach will help in the customer feedback workflow. "Areas" in the workflow represent the main topics while "Issues" refer to the sub-topics drilling down the main topics. The below example is just a conceptual example. Given the feedback regarding the Microsoft Outlook email application, the first component in our system can detect the main topic as Calendar through supervised learning. Yet there are so many features and potential issues about Calendar, e.g. change an appointment, create an event, add holidays to the calendar, share calendar, etc. The second component in our system is here to help identifying the specific sub-topic under the main topic through unsupervised learning.

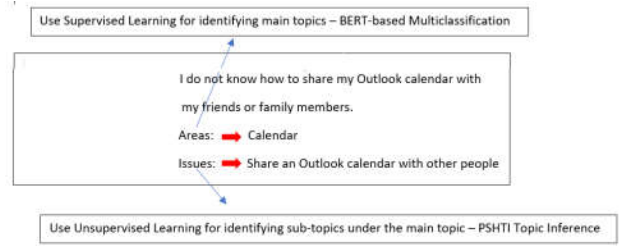

Fig. 1. An example how the two components in this work identify the main topic and the sub-topic for a given feedback

The existing unsupervised learning-based topic modeling or clustering methods usually select top words or terms as the primitive labels [8-11]. Yet, in practice, it is difficult to interpret a coherent representation or meaning of a topic given those selected top words. It requires a lot of manual work to convert the top words to more understandable labels. The second component in our new system enables the automatic mapping from the top words to the self-help issues with the help of domain knowledge extraction.

This work is not only novel in the methodology perspective but also influential in the real business and production. It prioritizes engineering investments in customer facing self-help solutions, as well as within the application and service. Moreover, it contributes to increased product quality by supplementing the product team to make data driven decision with regards to the feature requests, bug fixes, and architectural changes.

## II. METHODOLOGY

*A. Overview of the New System*

As mentioned earlier, we propose a two-component Semi-Supervised Learning system. Most cloud-based services and mobile applications have feedback from different channels, such as store reviews including Windows store reviews, Mac store reviews, iTunes store reviews and Google Play reviews, customer problem descriptions, help content and product feedback, product sentiment including NPS (Net Promoter Score) and surveys. This system gathers raw data from all channels and performs preprocessing.

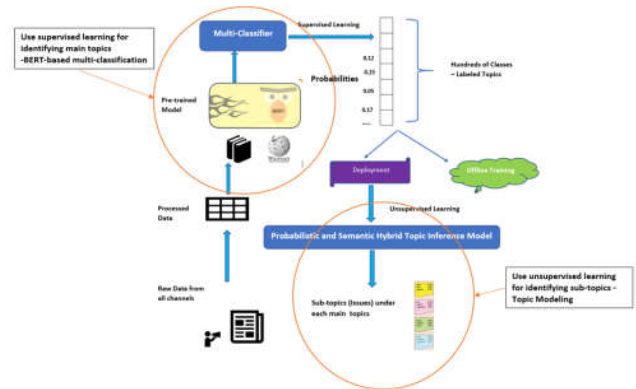

Fig. 2. The two-component Semi-Supervised Learning System

The first component – BERT-based multiclassification model aims at predicting the main topics given the input feedback among the hundreds of targeted topic classes. We have offline-training and online-serving, since this is a supervised learning way. Details will be explained in the following session B.

The second component is a supplement of the first component which incorporates both the state-of-the-art deep learning and external domain knowledge into the traditional topic modeling. It utilizes an unsupervised learning approach to infer the sub-topics given predicted main topics. Details will be found in the following section C.

### B. BERT-based Multiclassification Model to Identify Main Topics

In the offline-training for the BERT-based multiclassification component, we build a feed-forward classification neural network with Softmax on top of BERT (We used BERT Base in our system). Transfer learning is employed to fine-tune the parameters in both BERT and the Feed-forward Neural Network through training. Fig 3 illustrates the architecture of this component and shows how feedback will be fed into the system using a very simple example.

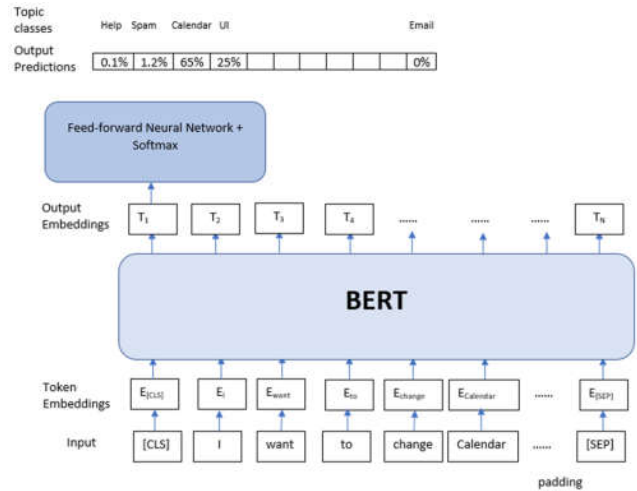

Fig. 3. Example how one simple feedback goes through our BERT-based Multiclassification Model

Specifically speaking, given the example feedback "I want to change Calendar", the beginning token is supplied with a special token [CLS] which stands for classification task while the ending token is marked as token [SEP] which represents the end of the feedback input. The token embeddings represents the word vector for each token and will be added by the positional embeddings when being fed into BERT. BERT is essentially a Transformer Encoder stack which consists of multiple encoders of self-attention layer and feed-forward neural network layer [1]. In practice, we have feedback in different lengths, so we have to set a standard maximum length for the input feedback where we need to do padding for the shorter feedback. In our experiment, we set the maximum length as 192, since we observed that most feedback have shorter than 192 words or the key user intents were explained within the first 192 words in most cases. The multi-classification model is trained offline against the labeled feedback data through batch gradient descent with cross-entropy as the target topic classes. Certainly, in the experiment, we may adjust your maximum length according to the characteristic of the textual data.

In the online-rendering, given every input feedback, our trained multiclassification model predicts the probabilities for all the target topic classes. For the same feedback example in Fig3, suppose we have target topic classes including "Help", "Spam", "Calendar", "UI" and so on. The multiclassification model outputs the probabilities for all the target classes given the input feedback "I want to change Calendar". The class "calendar" has the highest probability 65%, and the class "UI" has the second highest probability 25%.

### C. Probabilistic and Semantic Hybrid Topic Inference Model (PSHTI) to Identify Sub-topics

As the traditional topic modeling or clustering misses out the benefit from the trending deep learning, there have been some efforts in leveraging the semantic representation from deep learning into topic models since early this year [11]. Similar to the work in [11], our system is also to incorporate the topic probabilistic information from LDA-based topic model and semantic information from a pre-trained Sentence-BERT model [3] to learn a better representation of the textual feedback. However, our system is more advanced in the following perspectives: 1). Our PSHTI model utilizes two stages: firstly, it generates top-10 words for each sub-topic cluster under a main topic, secondly, it automatically maps top-10 words to self-help issues using domain knowledge through web crawling; 2). To concatenate the probabilistic embeddings and semantic embeddings, we carry out vector normalization for the embeddings generated from SBERT [3] to convert the elements in the vector between 0 and 1; 3). It provides more flexibility for the trade-off between LDA and SBERT by adding a prior parameter. More explanations are included in C.1.

From the BERT-based Multiclassification model, the corpus of the massive input feedback is predicted into different main topic classes (The same feedback might be included in multiple topic classes depending on your setup). For each main topic class, the group of feedback within that particular main topic is fed into PSHTI model.

### C.1. First Stage in PSHTI Model to Generate Top 10 Words for Each Sub-topic Cluster

As mentioned earlier, the first stage in PSHTI model shown in Fig 4 combines both the probabilistic information and the semantic information. Suppose we pre-define the number of sub-topic clusters as k given a certain main topic, then for the ith feedback in the group, we can get k-dimension vector where each number represents the probability of each topic cluster through LDA-based topic modeling: $A^i = [a_{i1}, a_{i2}, a_{i3}, ……, a_{ik}]$, e.g. $A^i = [0.1, 0.1, 0.1, 0.1, 0.2, 0.2, 0.1, 0.1]$ for 8 topic clusters. In contrast, for the ith feedback, Sentence-BERT produces 768-dimension vector $B^i = [b_{i1}, b_{i2}, b_{i3}, ……, a_{i768}]$, e.g. $B^i = [-0.2, -0.8, 1.3, ……, 2.4]$. To concatenate $A^i$ and $B^i$, we follow the steps as below:

- Normalize the elements in vector $B^i$ to make the numbers lay within the range between 0 and 1, since all the elements in vector $A^i$ are between 0 and 1. Suppose max_B and min_B are respectively the maximum number and the minimum number among the feedback corpus for the specific main topic $B^1, B^2, ……, B^N$. The normalization function for each element $b_{ij}$ is $(b_{ij} - min\_B)/(max\_B - min\_B)$. The normalized new vector is $B^i\_new$.

- Multiply each element in $A^i$ with the prior parameter gamma to get $A^i\_new$. The purpose is to give more weights to LDA-based topic modeling to keep more probabilistic information. gamma can be tuned.

- Finally concatenate $A^i\_new$ and $B^i\_new$ to get a 776-dimension vector.

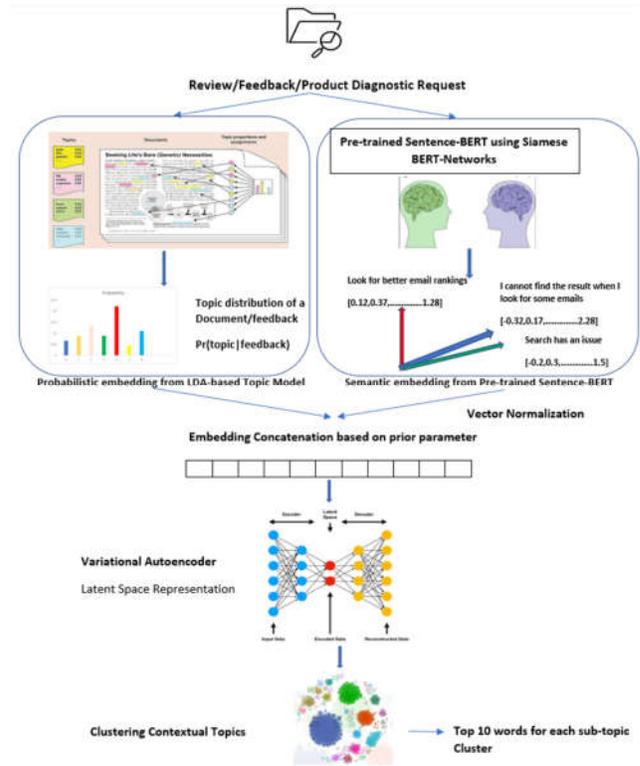

Fig. 4. The architecture of the first-stage in PSHTI model

The main purpose of Variational autoencoder is to learn a lower dimensional latent space representation of the concatenated vector generated from the above steps with the help of an unsupervised-learning way. Then, k-mean clustering was applied to cluster contextual topics and primitively output top-10 words for each sub-topic cluster.

### C.2. Second Stage in PSHTI Model to Generate Meaningful Issue Labels from Top-10-word Sub-topics

The existing work mostly produces top terms or require some post-processing to manually create more meaningful labels. However, our system enables the automatic labeling from the top words by incorporating the domain knowledge. Fig 5 displays the overall workflow with a simple example.

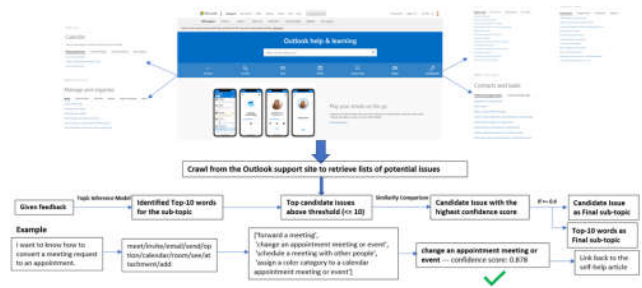

Fig. 5. The architecture of the second-stage in PSHTI model with a simple example

Take the feedback for Microsoft Outlook as an example, we web-crawl from the customer support site to retrieve a domain knowledge library of the potential issues in all perspectives: Calendar, manage and organize, email, customize, contacts, tasks and so on. Each issue is linked to some self-help content. For instance, we have feedback "I want to know how to convert a meeting request to an appointment", "Calendar" is predicted as one of the main topics from the BERT-based Multiclassification model. Within "Calendar" main topic, this feedback and all the other feedback related to "Calendar" is fed into the first stage in PSHTI topic inference model. It is clustered to the sub-topic "meet/invite/email/send/option/calendar/room/see/attachment/add". In the second stage, the cosine similarity score between the Sentence-BERT embedding vector of this clustered sub-topic and the Sentence-BERT embedding vector of each issue in the domain knowledge library will be computed. The top 10 issue candidates with the highest score above some threshold are generated. In many cases, there might not be a candidate due to the incompleteness of the issue list or no similar issues found. In this example shown in Fig 5, we have five candidates: "forward a meeting", "change an appointment meeting or event", "schedule a meeting with other people", "assign a color category to a calendar appointment meeting or event". These five candidates are then compared with the original feedback based on Sentence-BERT embedding vectors. The final confidence score is the similarity score between the candidate embedding and the original feedback embedding. If the confidence score of the highest candidate issue is no lower than 0.6, the candidate issue will be taken as the final sub-topic. In our example, the final topic "change an appointment meeting or event" achieves the highest score 0.878. As a result, in the real production, when a customer submits feedback in the client application or files a support request in the diagnostics system, our system automatically detects a self-help solution with content. However, in some cases, we still do not have a candidate from the domain knowledge library where the top 10-words are still used as the final sub-topic. In the further work, we need enlarge our domain knowledge library and improve the system.

III. EXPERIMENT AND EVALUATION

*A. Data*

We experiment on feedback data from January 2019 to March 2020 with 102 labeled main topics. After removing short, un-translated and noisy texts, we get about 1.3 million translated feedback from different channels. Moreover, we perform random sampling and get around 1.3 million feedback-main topic pairs for training and around 210,000 pairs for testing. Around 70% feedback have multiple main topics. Certainly, no feedback in the training data should show up in the testing set, so we needed to perform random sampling on unique feedback items.

We also performed some basic study about the data. Fig 6 illustrates the distribution of the number of identified main topics per unique feedback for training and testing data. Around 70% feedback have multiple main topics. Moreover, we also studied the distributions of all the 102 labeled main topics, and the top 20 are shown in Fig 7. The top 5 topics include Help Content, UI/UX/Add-in, Account/Authentication, Calendar and Send/Receive.

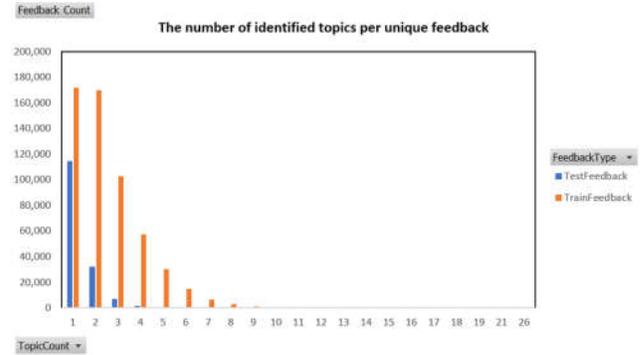

Fig. 6. The number of main topics per unique feedback respectively in the training and testing data set

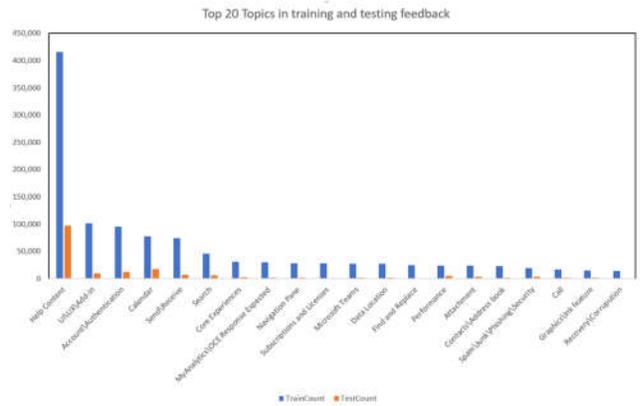

Fig. 7. Top 20 Topics in training and testing feedback

*B. Experiment and Evaluation Metrics*

We trained the first component – BERT-based Multiclassification model based on the training data set mentioned in A and tested on the testing data set using accuracy@Top 1, Top2 and Top 3 as the evaluation metric. Since most feedback has multiple main topics, we decided to use Accuracy@Top 3 as the final evaluation metric and predict 3 top main topics for the new upcoming feedback in production. Moreover, there used to be some noises, html style texts in the feedback, we compared the evaluation results between the trained models before and after cleaning, shown in Table I. The

accuracy@Top 3 after cleaning the data is 0.91. In another word, if we automatically classify each of the upcoming 100 feedback to the three most likely topics, 91 feedback would be correctly classified. This is 11% increase from before cleaning the data. It is concluded that the model is robust even if the data is a bit noisy.

TABLE I. EVALUATION RESULTS OF BERT-BASED MULTICLASSIFICATION MODEL

| Evaluation Results | | |
|---|---|---|
| **Evaluation Metric** | **Model trained on noisy data** | **Model trained on clean data** |
| Accuracy@Top 1 | 0.65 | 0.66 |
| Accuracy@Top 2 | 0.84 | 0.85 |
| Accuracy@Top 3 | **0.9** | **0.91** |

We tested the second component – PSHTI topic inference model based on the testing data. There are 82 predicted main topics in the testing data, and we ran PSHTI for each main topic group where the number of sub-topic clusters were set to be 8 for the group with more than 500 feedback and 2 for the rest. We chose UCI Coherence Score and Silhouette Score as the evaluation metrics. UCI Coherence Score is a slightly improved version of UMass, and applying the sliding window based probabilistic measure [12, 13]. According to the paper [13], UCI Coherence Score is discovered to be most correlated to human ratings among all the topic coherence evaluation metrics. It measures how much the top-10 words of a topic are related each other. The equation is:

$$C_{UCI} = \frac{2}{N*(N-1)} \sum_{i=1}^{N-1} \sum_{j=i+1}^{N} PMI(w_i + w_j)$$

$$PMI(w_i + w_j) = \log \frac{P(w_i, w_j) + e}{P(w_i) * P(w_j)}$$

N is the number of the top words from a Topic, here is defined as 10. $P(w_i, w_j)$ is the probability of the co-occurrence between the two words $w_i$ and $w_j$, where the co−occurrence is calculated by applying the sliding window over the text document [13]. Silhouette Score is used to determine the degree of separation between the clusters [14, 15] for the clustering algorithms. For each sample, the calculation includes the following steps:

- Compute the average distance a(i) from all the data points in the same cluster;
- Compute the average distance b(i) from all the data points in the closest cluster;
- Compute the silhouette of the given sample: $s(i) = \frac{b(i)-a(i)}{\max\{a(i),b(i)\}}$

The Silhouette Score is the mean Silhouette Coefficient over all the samples [16].

Table II displays the evaluation results for LDA based topic modeling, BERT-based clustering, our model PSHTI with normalization and without normalization. Our Model PSHTI with normalization outer-performs other methods, especially the pure semantic based clustering and pure probabilistic based topic modeling.

TABLE II. EVALUATION RESULTS OF PSHTI MODEL

| Model | Silhouette Score | Coherence Score |
|---|---|---|
| **PSHTI Inference Model with normalization** | **0.59** | **0.392** |
| **PSHTI Inference Model without normalization** | *0.32* | *0.385* |
| **LDA** | *0.55* | *0.377* |
| **BERT-based clustering** | 0.06 | *0.391* |

IV. CONCLUSION

In this paper, we proposed a novel two-component Semi-supervised learning system. For a given feedback or support request, the first component – BERT-based multiclassification model can predict the top main topics, while the second component – PSHTI model can further identify the sub-topics under the predicted main topics. We evaluated the performance of both components and observed that both components achieved great results. Moreover, the second component takes the traditional topic model or topic categorization methods to the next level by incorporating the domain knowledge to the automatic generation of meaningful topic labels or self-help topics.

From a business perspective, the results from the second PSHTI component help focus engineering investments. The ability to see detailed areas of feedback instead of just a high-level topic area can provide greater insight into problem areas, bugs, diagnostic tools and help documentation. In addition, the greater granularity can inform product training, deployment issues, support engineer training and helps inform almost every aspect of the business. With large volumes, this information is impossible to get at scale. Certainly customer feedback, public comments, blogs, and support information can be reviewed and consumed on a piecemeal basis, but this two component process provides a level of detail and aggregation that is not possible to accomplish manually.

Future work will include more investigations on the customized number of sub-topic clusters within a given main topics. We will continue to tune domain knowledge by adding new sources and engineering feedback. As product and service features evolve, this will be an ever-changing set of results that will allow us to constantly improve the model. We also would like to explore the relationships between sub-topic output and service performance, and explore the creation of predictive models and alerting mechanisms.

ACKNOWLEDGMENTS

Completion of this work could not have been possible without the collaboration of many teams across Microsoft and the industry. We have a special callout to the Microsoft feedback team for enabling us to leverage their already established feedback repository and the Outlook product group for graciously letting us implement the model in that specific area.

We would like to personally thank Jonas Gunnemo, Ricardo Stern, Huw Upshell, Danylo Fitel, Scott Carter, Yan Guo, and Jason Bluming for their support and collaboration.

Finally, we would like to acknowledge the many different Microsoft teams who shared their help and guidance on privacy and security.